\documentclass[10pt,twocolumn,letterpaper]{article}

\usepackage[pagenumbers]{cvpr} 

\usepackage[dvipsnames]{xcolor}

\usepackage{placeins}

\definecolor{cvprblue}{rgb}{0.21,0.49,0.74}
\usepackage[pagebackref,breaklinks,colorlinks,citecolor=cvprblue]{hyperref}

\def\paperID{*****} 
\def\confName{CVPR}
\def\confYear{2024}

\title{Radial Networks: Dynamic Layer Routing \\ for High-Performance Large Language Models}

\author{
    Jordan Dotzel\thanks{Equal Contribution} \qquad Yash Akhauri\footnotemark[1] \qquad Ahmed S. AbouElhamayed \qquad Carly Jiang\\
    Mohamed Abdelfattah \qquad Zhiru Zhang\\
    Cornell University\\
    {\tt\small \{dotzel, ya255, afa55, cjj43, mohamed, zhiruz\}@cornell.edu}
}

\begin{document}
\maketitle
\begin{abstract}
Large language models (LLMs) often struggle with strict memory, latency, and power demands.
To meet these demands, various forms of dynamic sparsity have been proposed that reduce compute on an input-by-input basis.
These methods improve over static methods by exploiting the variance across individual inputs, which has steadily grown with the exponential increase in training data.
Yet, the increasing depth within modern models, currently with hundreds of layers, has opened opportunities for dynamic \textit{layer sparsity}, which skips the computation for entire layers.
In this work, we explore the practicality of layer sparsity  by profiling residual connections and establish the relationship between model depth and layer sparsity.
For example, the residual blocks in the OPT-66B model have a median contribution of 5\% to its output.
We then take advantage of this dynamic sparsity and  propose \textbf{Radial Networks}, which perform token-level routing between layers guided by a trained router module.
These networks can be used in a post-training distillation from sequential networks or trained from scratch to co-learn the router and layer weights.
They enable scaling to larger model sizes by decoupling the number of layers from the dynamic depth of the network, and their design allows for layer reuse.
By varying the compute token by token, they reduce the overall resources needed for generating entire sequences.
Overall, this leads to larger capacity networks with significantly lower compute and serving costs.

\end{abstract}    
\vspace{-5pt}
\section{Introduction}
\label{sec:intro}

\begin{figure}[t]
  \centering
  \includegraphics[width=.7\linewidth]{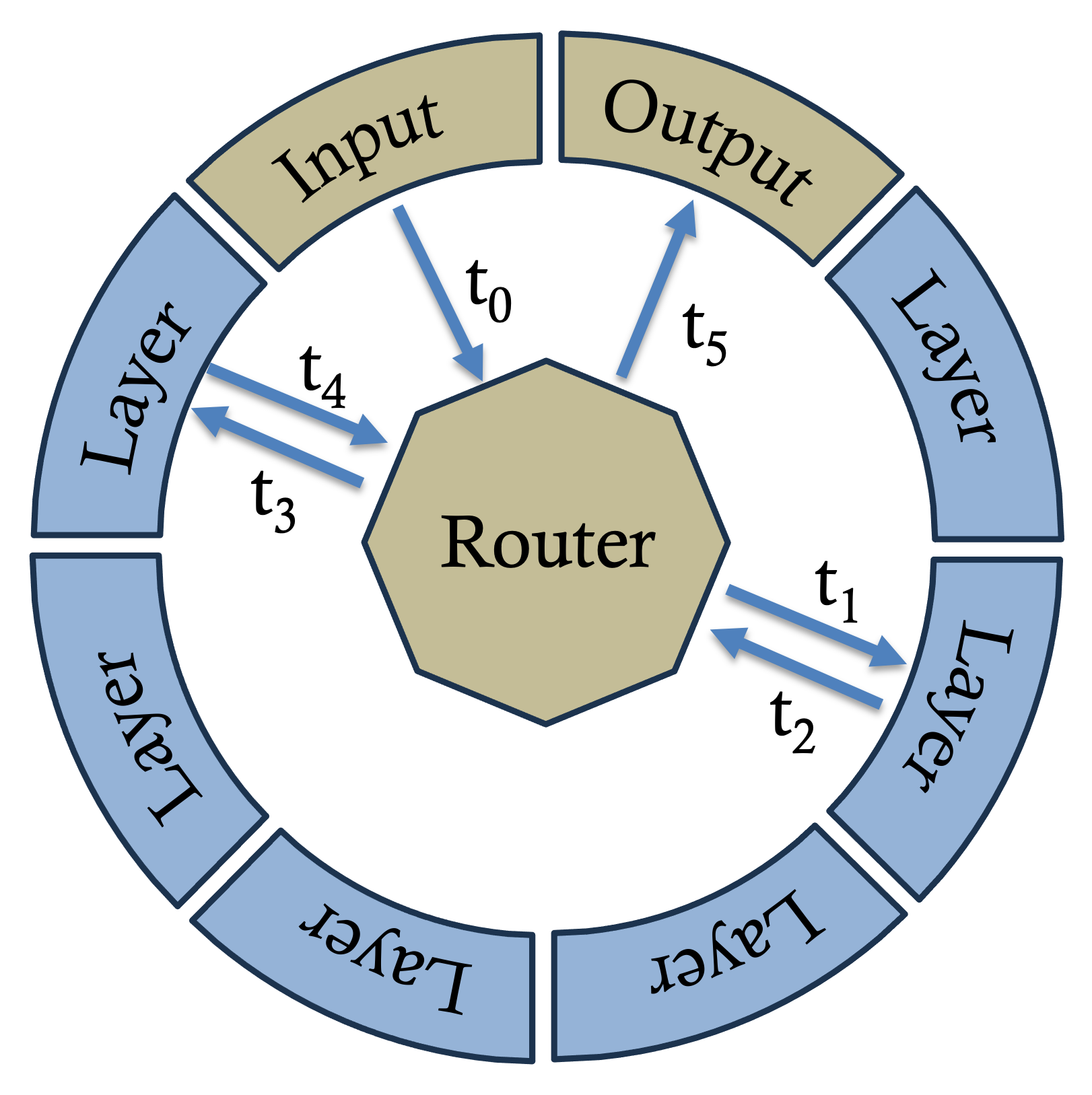}
      \caption{\textbf{Radial Networks --} Radial Networks generalize sequential networks for higher accuracy and performance. They take advantage of significant dynamic layer sparsity within modern LLMs and invoke only a  subset of model layers for each token. They reduce the average network depth, lower model latency, and provide a more scalable neural architecture.}
  \vspace{-5pt}
  \label{fig:radial}
\end{figure}

\begin{figure}[t]
  \centering
  \includegraphics[width=\linewidth]{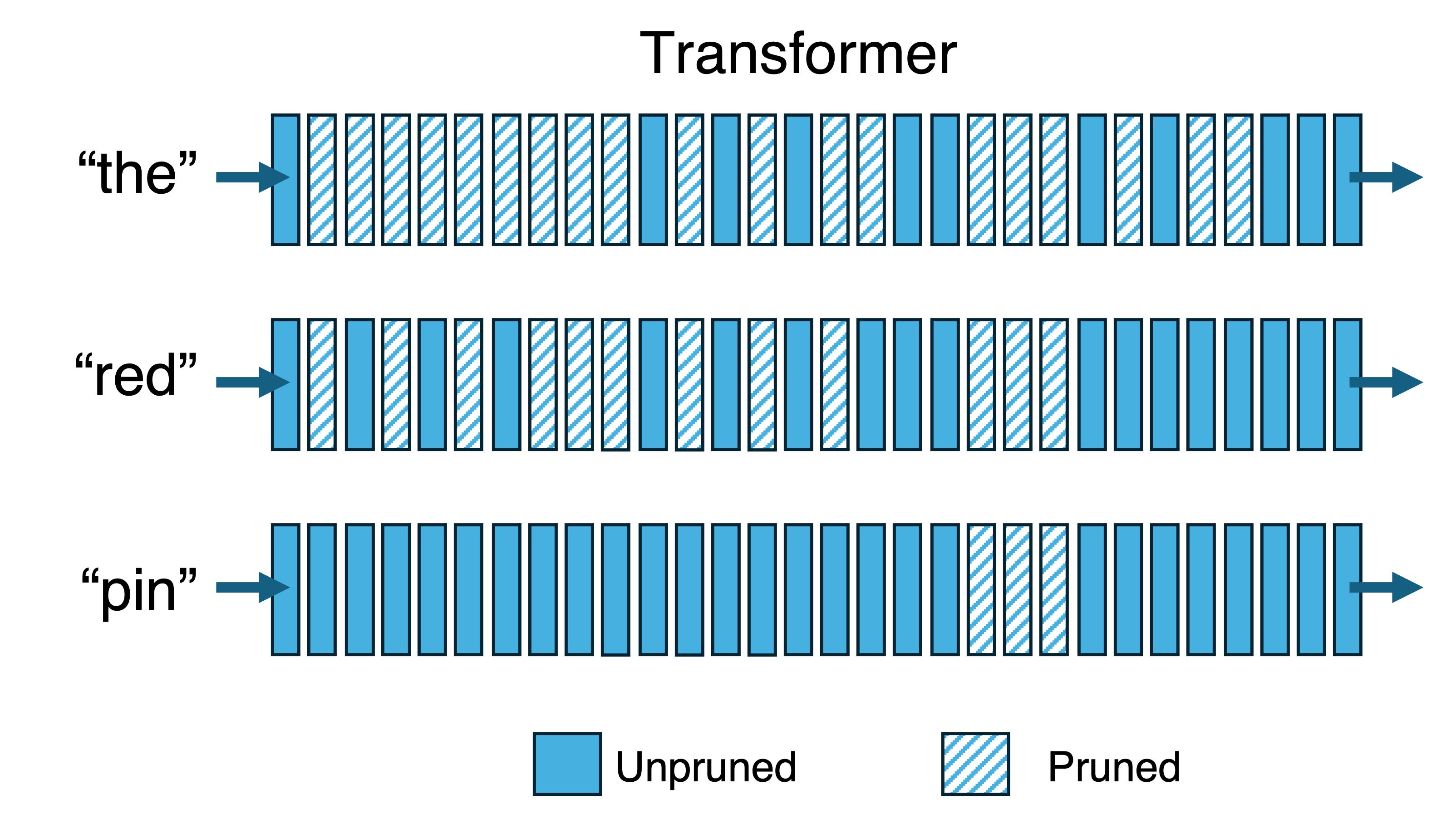}
    \caption{\textbf{Dynamic Layer Sparsity --} As transformers grow larger, each layer contributes less to the output and shows significant variation on a token by token basis. Dynamically pruning these layers allows for models to grow significantly without corresponding increases in model latency. Early profiling results suggest that individual layers contribute around 1\% within modern state-of-the-art language models. }
    \vspace{-5pt}
  \label{fig:tokens}
\end{figure}

Large language and vision models have recently achieved state-of-the-art performance across various tasks, yet due to their large computational requirements, they struggle with strict memory, latency, and power demands.
As these transformers grow larger, they create opportunities for dynamic layer sparsity, which can skip individual layers on an input-by-input basis, as shown in Figure~\ref{fig:tokens}.
For instance, our residual block profiling in Section~\ref{sec:profiling} suggests that modern state-of-the-art transformers likely have a median contribution around 1\% to the output at each block, and that these contributions are \textit{dynamic}, varying token by token.
This type of sparsity was impractical at smaller scales and with previous neural architectures.
At smaller scales, every layer contributes significantly to the computation for each input, and with previous architectures, e.g., convolutional neural networks (CNNs), models change their intermediate dimensions throughout their depth and skipping causes dimensional mismatches.

This work shows that the layer contributions vary among models and tasks, and often the earlier layers of the network contribute more than the later layers.
This indicates that early-exit methods, which dynamically prune the later layers in the network, often focus on the wrong set of layers.
This dynamic contribution can be exploited at the token-level if it can be predicted accurately and efficiently at runtime. 
This work explores the opportunities for dynamic sparsity within the modern transformers by focusing on the OPT family of models~\cite{zhang2022opt} for language and ViT models~\cite{dosovitskiy2021image} for vision.
It profiles the residual blocks to quantify the importance of each intermediate layer to its output and then highlights trends across model size and block types.
Then, it inserts \textit{oracles} at every layer to calculate various accuracy proxies and simulate greedy decisions on which layers to dynamically skip per token.
\vspace{-5pt}
\section{Related Work}
\label{sec:related}

\begin{figure}[t]
  \centering
  \includegraphics[width=\linewidth]{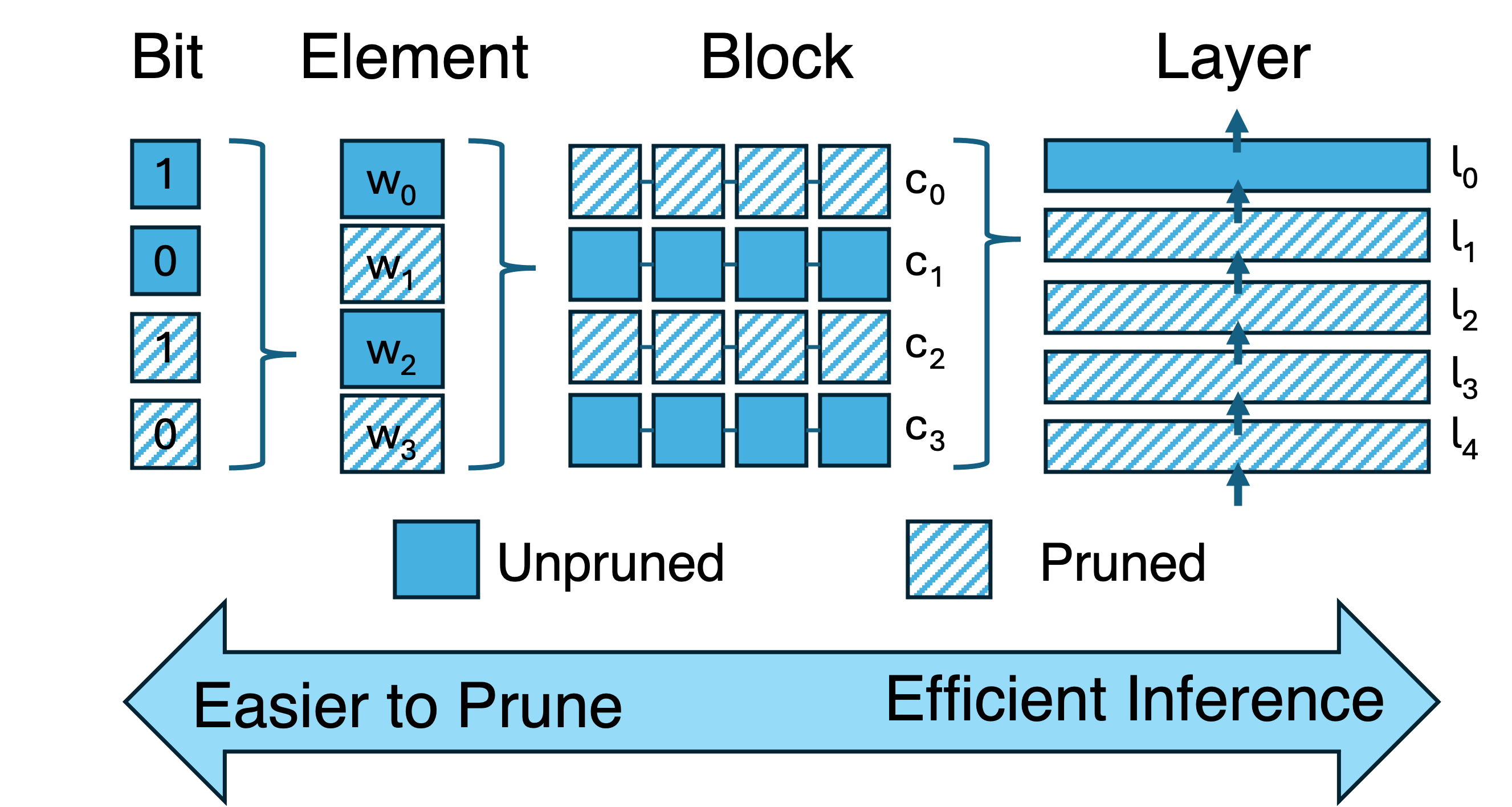}
      \caption{\textbf{Sparsity Granularity --} Bits form the basis for elements (weight or activations), which create blocks (rows, columns, heads), which then form individual layers. This leads to a sparsity spectrum where the smaller units are easier to prune without accuracy loss yet more difficult to accelerate. Layer sparsity has the highest potential for inference speedup and largest support within current hardware.}
  \vspace{-5pt}
  \label{fig:sparsity}
\end{figure}

Sparsity research with deep neural networks has a long history, and broadly can be categorized in terms of granularity, structure, and mode (static vs. dynamic)~\cite{hao2021codesign}.
Figure~\ref{fig:sparsity} shows sparsity granularity, beginning with bits that construct parameter elements, elements that build blocks, and blocks that form layers.
As the unit becomes larger, it becomes more difficult to arbitrarily prune without accuracy loss yet easier to accelerate with modern hardware.
For instance, unstructured element sparsity in weights leads to high compression levels while maintaining model accuracy, yet it requires specializing sparse accelerators to translate compression into end-to-end speedup.

In addition, the sparsity mode can either be static or dynamic.
Static sparsity leads to more regular patterns that can be optimized by compilers and simpler architecture that do not need additional sparsity predictors, yet it must apply to all inputs together.
In contrast, dynamic sparsity can take advantage of input-dependent characteristics to increase model accuracy at higher levels of compression.
This work focuses on dynamic layer sparsity, which can take advantage of the recent explosion in model depth within language models.

\subsection{Dynamic Sparsity}
\label{sec:dynamic}

Multiple prior works have proposed dynamic sparsity to accelerate DNNs across granularities.
For example, Channel Gating introduced a method for dynamic channel sparsity that reduced the compute of CNN workloads by up to $8\times$ without significant accuracy loss~\cite{hua2019channel}.
Precision Gating continued this line of research by applying dynamic sparsity at the bit level to reduce the required compute by up to $3\times$~\cite{zhang2020pg}.
Later, DejaVu applied a similar approach within LLMs to induce dynamic sparsity on the channels within the FFN layer and across the heads of the attention layer~\cite{liu2023deja}.

\subsection{Early Exit}
\label{sec:early}

In addition to dynamic sparsity along the network width, multiple prior works have explored sparsity in the depth dimension.
For instance, early-exit DNNs use dynamic sparsity along the depth dimension by allowing the computation to exit prematurely at fixed points within the network~\cite{teera2016early, bolukbasi2017adaptive}.
This process must be trained end-to-end using a joint loss function that weights the contributions from each early-exit layer.
However, this work shows that in many models, the earlier layers in the model often contribute more, and therefore early-exits are significantly more difficult to apply post-training.

\vspace{-5pt}
\section{Layer Sparsity}
\label{sec:sparsity}

\begin{figure}[t]
  \centering
  \includegraphics[width=\linewidth]{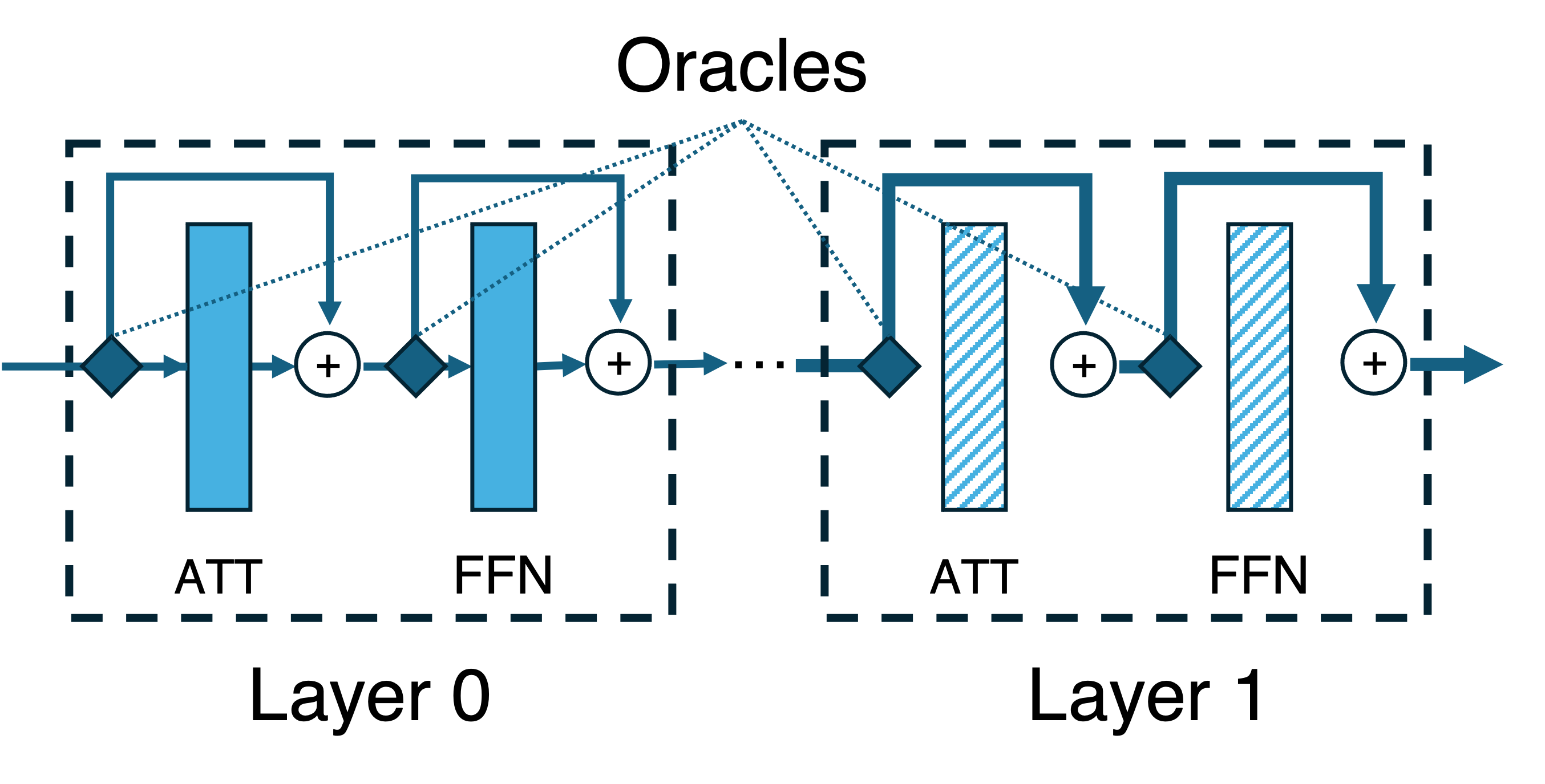}
      \caption{\textbf{Residual Blocks --} There are two types of residual blocks within transformers, attention (ATT) and feed-forward network (FFN). These blocks offer natural points to profile layer strength since block inputs and outputs are combined at a single point. To establish an upper bound on the effectiveness of dynamic layer sparsity, oracles are inserted before each block that know the layer contribution beforehand.}
  \vspace{-5pt}
  \label{fig:residual-skip}
\end{figure}

Transformer layers contain two residual blocks: attention (ATT) and feed-forward network (FFN)~\cite{vaswani2017attention}.
These blocks each contain the main residual branch $R(x)$, which comprises multiple individual layers, and the identity branch $x$, which bypasses the residual branch and simply returns its input.
They combine these branch outputs together to compute $R(x) + x$, so that during training the main branch only has to learn the function residual $R(x) - x$.

These blocks offer natural breakpoints within the model to profile and induce layer sparsity since they already provide skip-connections that have been trained along with the model.
Figure~\ref{fig:residual-skip} shows a lower-level view of these blocks within two transformer layers.
It shows that the main branch and skip connections are combined at an addition node before they are passed to the next block.
This structure enables easy profiling of the blocks by measuring the relative magnitudes into these additions.

This figure also shows the insertion of oracles that can switch on and off the main branch using various accuracy proxies, such as the residual ratio as defined in Section~\ref{sec:residual}.
When they are switched on, the block operates normally combining the skip and residual branches, and when switched off, only the skip connection is active.
This work focuses on the opportunities for dynamic layer sparsity and simulates layer skipping by allowing these oracles to have access to future information.
\section{Profiling}
\label{sec:profiling}

The primary proxy used by these oracles is the \textit{residual ratio}, which captures the relative importance of the main and skip branches. 
This section uses this ratio to analyze the layer sparsity within OPT and ViT models with examples taken from WikiText-2 and COCO.
The WikiText-2 examples are packed together to avoid the use of padding to simulate batch-size one inference.
This batch-size one setting is very common in practice and avoids many complications with dynamic layer sparsity that arise when using batches of examples.

\subsection{Residual Ratio}
\label{sec:residual}

To profile these opportunities for dynamic sparsity, this section defines the residual ratio $r$ as:

\vspace{-8pt}
\begin{equation}
\text{r} = \frac{\|R(x)\|_2}{\|x\|_2}
\end{equation}
\vspace{-8pt}

This simple quantity captures the contribution of the residual branch, and acts as an efficient post-training proxy for more expensive metrics, such as empirical layer sensitivities.
For example, a block with a 2\% 
residual ratio indicates the main branch provides a 2\% average contribution at the output, although there can be large element-wise variance.
Therefore, skipping blocks with ratios this small should have little overall effect on the output of the network.

\subsection{Model Size}
\label{sec:size}

\begin{figure}[t]
  \centering
  \includegraphics[width=\linewidth]{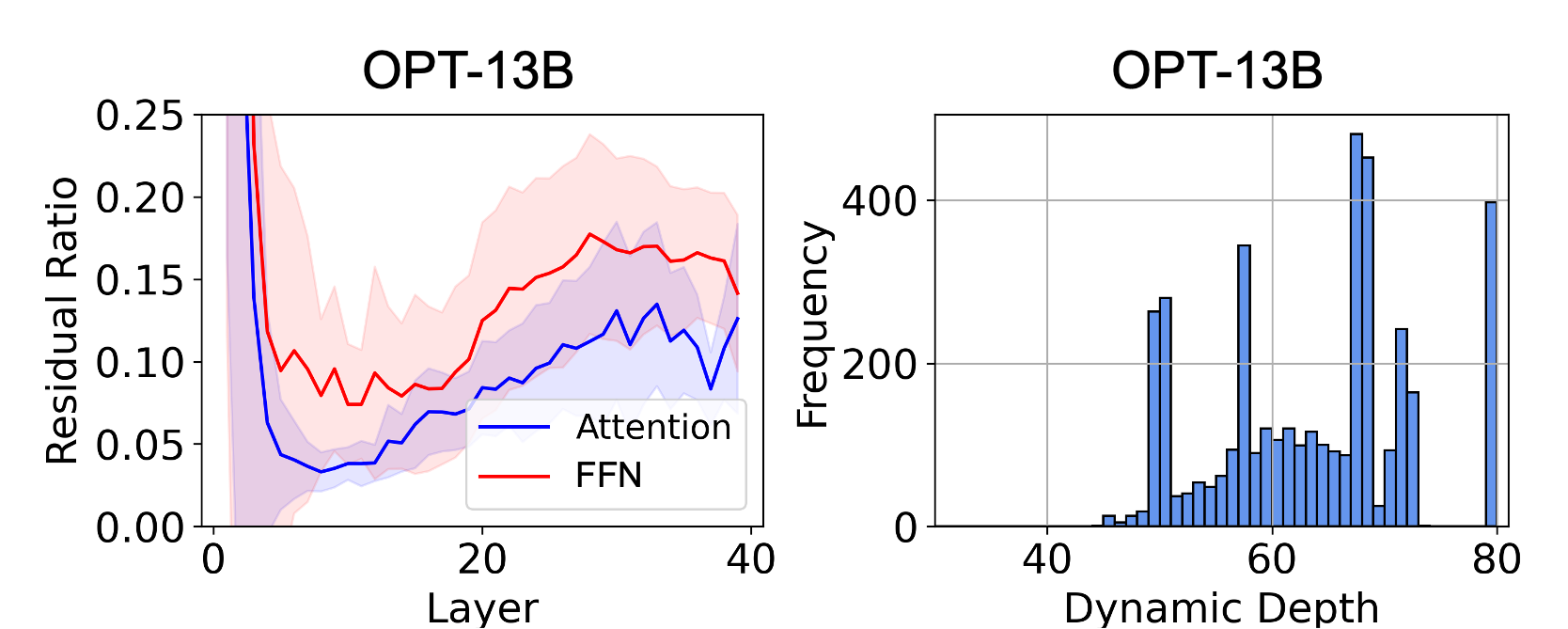}
      \caption{\textbf{Dynamic Depth --} The deeper layers in the network contribute more than the earlier layers, except for the very first layers. This relationship benefits from a routed architecture as opposed to early-exit, since early-exit skips the deeper layers. In addition, there is significant variance in the dynamic depth of the model, allowing for token-specific sparsity.}
  \vspace{-5pt}
  \label{fig:depth}
\end{figure}

\begin{figure}[t]
  \centering
  \includegraphics[width=\linewidth]{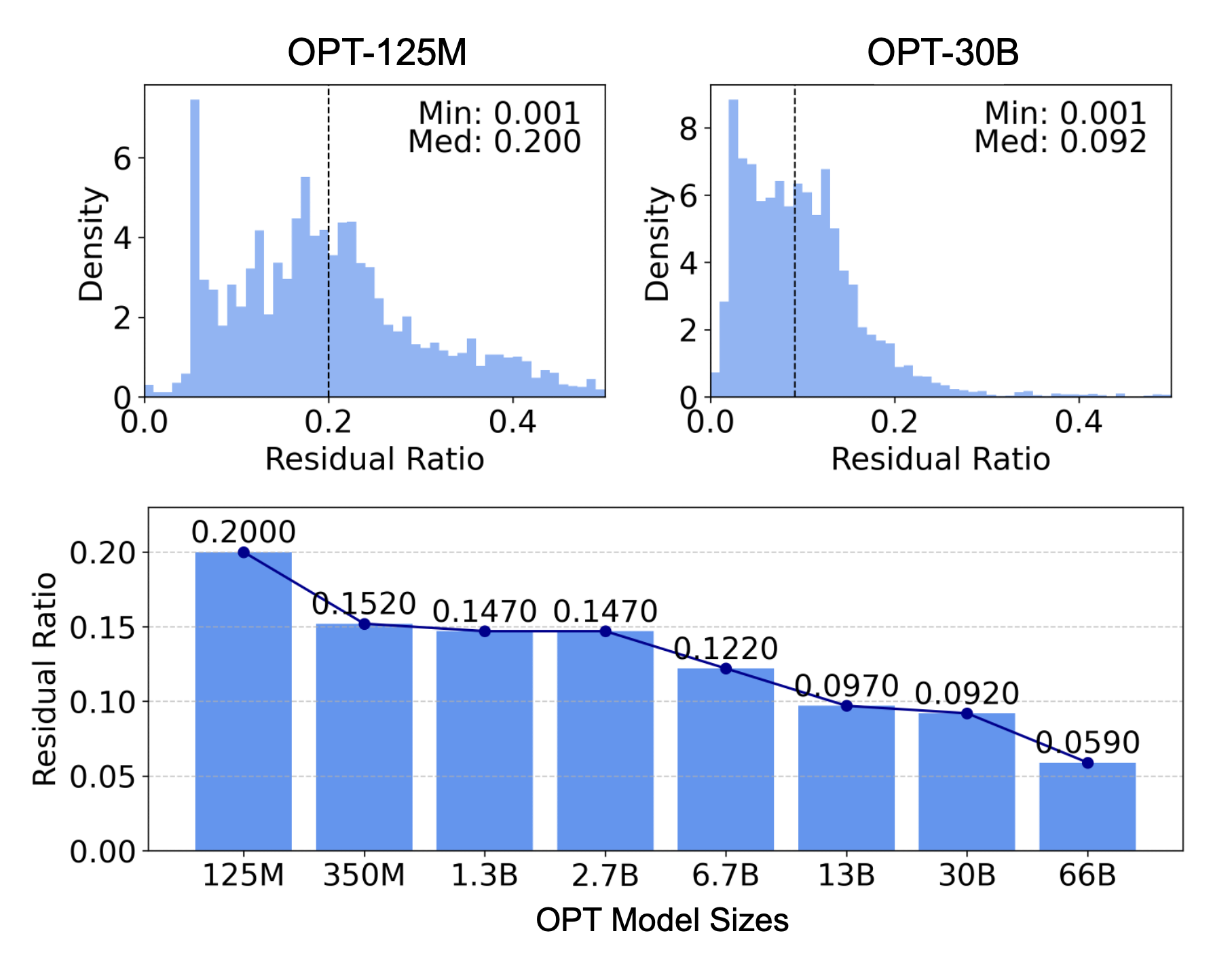}
      \caption{\textbf{Residual Ratio --} As models grow larger, the residual ratio decreases for each layer, and therefore more layers contribute less to the overall output. Each data point represents a residual ratio for a single token for both attention and feed-forward blocks.}
  \vspace{-5pt}
  \label{fig:pareto}
\end{figure}

\begin{figure}[t]
  \centering
  \includegraphics[width=\linewidth]{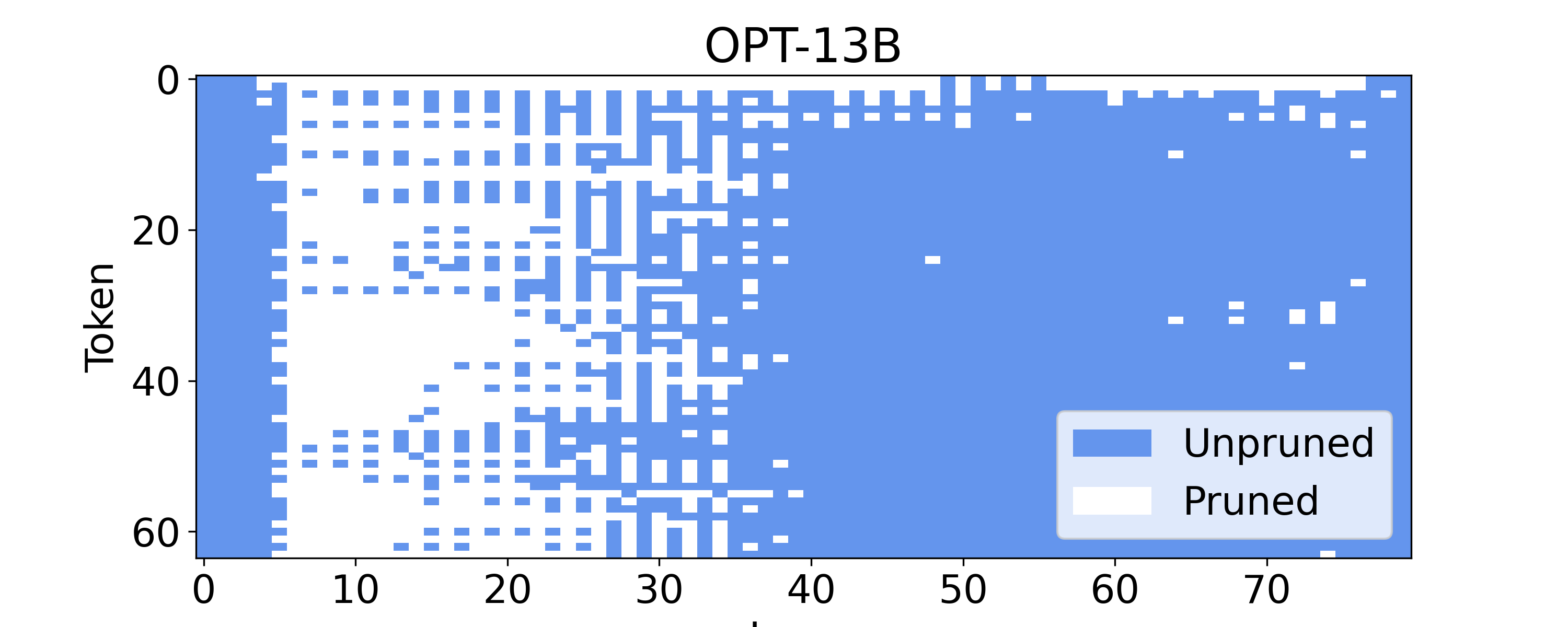}
    \vspace{-10pt}
      \caption{\textbf{OPT-13B Routing --} The majority of skipped layers are in the earlier part of the network. A layer skip is assumed when the residual ratio drops below 5\%. ATT and FFN layers are interleaved in the diagram. }
  \vspace{-5pt}
  \label{fig:opt-trace}
\end{figure}

This ratio can be used to understand the relationship between model size and dynamic sparsity.
Figure~\ref{fig:pareto} explores this by plotting the residual ratio across OPT models for the residual attention or feed-forward block (more plots shown in Appendix~\ref{sec:appendix}).
Each data point represents a single token during the model generation phase.
It shows that as the model size grows, the ratio distribution becomes more skewed to the left, indicating that opportunities for layer sparsity expand with model size. 
For instance, while the median residual ratio for OPT-125M is only 20\%, it drops to 5.9\% for OPT-66B.

In addition, the ratio seems to track the number of model parameters, not just the number of layers.
For example, OPT-2.7B and OPT-6.7B have the same number of layers, differing only in their hidden dimensions, yet the ratio for OPT-6.7B continues the decreasing trend.
This trend likely continues for even larger models, making dynamic layer sparsity more practical within modern state-of-the-art models with greater than one trillion parameters.

\subsection{Dynamic Depth}
\label{sec:depth}

Dynamic layer sparsity leads to dynamic depth networks that adjust their depth based on their model inputs.
Figure~\ref{fig:opt-layers} shows the residual ratio across the layers of an OPT-13B model (all models shown in Appendix~\ref{sec:appendix}).
All values shown are mean residual ratios taken across tokens from Wikitext-2 data using a sequence length of 256.
The ratio variance is highlighted in lighter colors centered around the mean.
It demonstrates that the earlier residual blocks contribute more compared to the later layers, except for the first few layers.
In addition, there is significant variance across tokens across layers suggesting the opportunity to apply dynamic layer sparsity to only the tokens with lower ratios.

This figure additionally shows the dynamic depth caused by this layer sparsity.
It assumes oracles that threshold the residual ratio at each block and skip the residual branch if it falls below this threshold.
Since computing the ratio requires running the residual branch, this is only used for profiling and simulation purposes.
Each data point represents an inference of a single token using a ratio threshold of 5\% is used.
The figure confirms a spread within the network depth, where most tokens only need between 40 and 70 blocks, instead of the full network at 80 blocks.
 
\subsection{Routing Traces}
\label{sec:trace}

For more detailed analysis, Figure~\ref{fig:opt-trace} shows the routing for the OPT-13B model across a batch of WikiText-2 examples.
It reveals how the lower residual ratios in Figure~\ref{fig:opt-layers} lead to a significant number of skipped layers in the beginning of the model.
This again motivates the use of dynamic layer sparsity over early-exit models, since early exit can only skip later layers, which contribute the most to the network.
 
\subsection{Vision}
\label{sec:vision}

\begin{figure}[t]
  \centering
  \includegraphics[width=\linewidth]{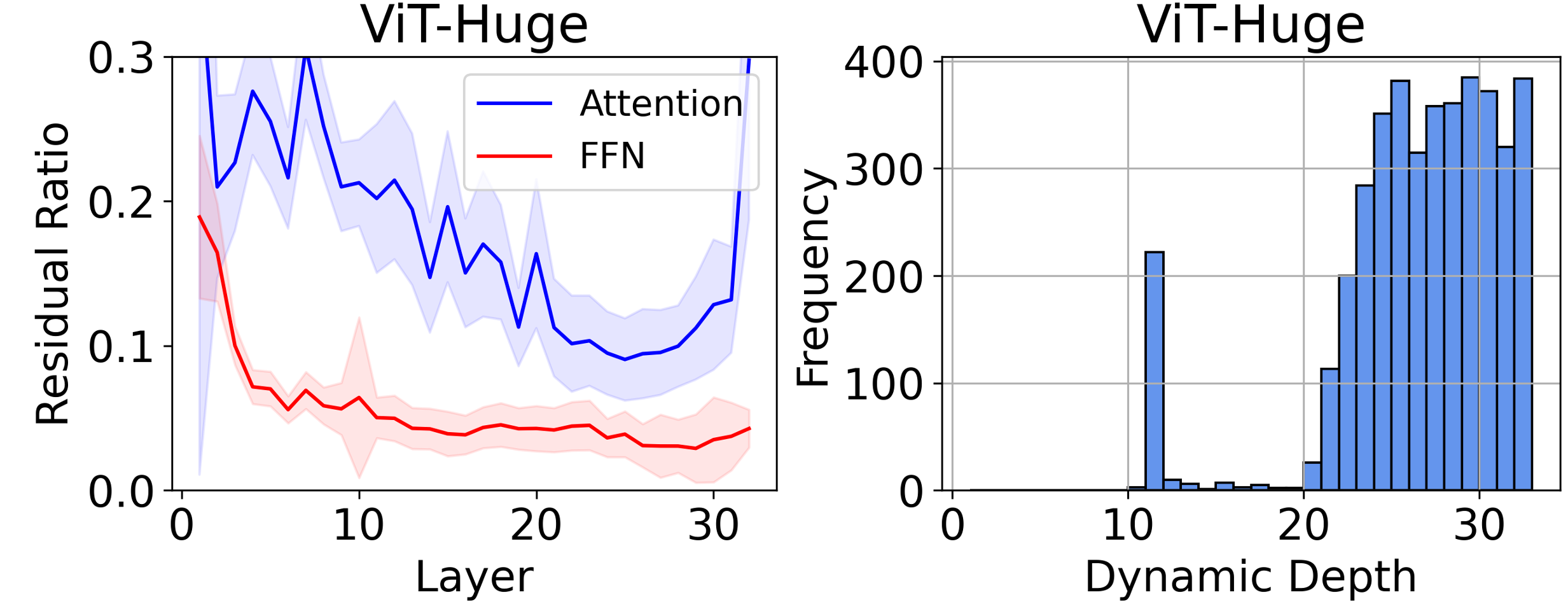}
    \vspace{-10pt}
    \caption{\textbf{Vision Profiling --}  Current vision transformers are substantially smaller than language models yet demonstrate similar trends with residual ratio and dynamic depth. ViT-Huge (632M) shows median ratios comparable to OPT-350M, and the dynamic depth assumes skipping layers with lower than 10\% residual ratio. }
    \vspace{-5pt}
  \label{fig:vit}
\end{figure}

This analysis so far has focused on large language models, since they are currently $10$ to $100\times$ larger than large vision models. 
Yet, recent vision transformers have been proposed with tens of billions of parameters~\cite{dehghani2023scaling}.
These weights are not yet released, yet the trends between the smaller language and vision models can still be aligned at smaller scales to suggest the behavior of large vision models with billions or trillions of parameters.

Figure~\ref{fig:vit} shows a comparison for the largest released ViT model, which contains 632M parameters across 24 layers.
It shows that vision transformers at this size have comparable residual ratios to the similarly sized OPT-350M.
In addition, Appendix~\ref{sec:appendix} lists smaller ViT versions and shows a similar trend between model size and residual ratio, suggesting that as vision transformers increase in size they will benefit from the same layer sparsity opportunities as the OPT models.
\section{Radial Networks}
\label{sec:radial}

Given the high degree of dynamic layer sparsity, we propose a new neural architecture that natively supports arbitrary routing between layers.
As shown in Fig~\ref{fig:radial}, each token enters network at step $t_0$ and then is routed dynamically to the next layer at later steps $t_n$.
This process allows for dynamic computation given the variable number of layers included in each token path.
As with standard transformers, the network continues auto-regressively until an end-of-sequence token is produced.

\subsection{Router}
\label{sec:router}

The router is the central component with the responsibility of directing each embedding between layers at each time step.
Building on the success of mixture-of-expert models, we learn this router with a small multi-layer perceptron (MLP) model.
Beginning from layer $l_t$, it maps from intermediate embeddings $e_t$ to output router logits $z_t$ for each layer.
These logits are then passed into a softmax function to produce probabilities $p_t$.
The maximum probability layer is then chosen as the next layer in the forward pass.

\vspace{-20pt}
\begin{align*}
    z_t &= Router(e_t) \\
    p_t &= \text{softmax}(z_t) \\
    l_{t+1} &= \text{argmax}(p_t) \\
\end{align*}
\vspace{-20pt}

The iterations stop when the model chooses the output layer, or a set maximum number of layers are seen, which forces the output layer.
This maximum number of layers is a hyper-parameter that limits the worst-case dynamic depth in the network.

\subsection{Unified Cache}
\label{sec:cache}

In standard sequential transformers, each layer activates for each token, and the attention mechanism references the key-value cache of previous tokens.
However, given the dynamic routing of radial networks, many of the layers are not activated for different tokens, and the key-value cache is sparse.
To solve this, we instead use shift the cache from the layers to a shared global cache that stores all of the previous key-value pairs.
Each embedding then attends to the cached pairs of previous iterations and previous tokens.

To distinguish between the cached values for the current token and previous, we use standard positional embeddings.
These embeddings are used to encode the give the relative position of tokens within the input sequence.
A common approach is to use sinusoidal functions; for position \( p \) and dimension \( i \), the positional embedding \( PE(p, i) \) can be defined as follows:

\[
PE(p, i) = 
\begin{cases} 
\sin(p / 10000^{i/d}) & \text{if } i \text{ is even} \\
\cos(p / 10000^{(i-1)/d}) & \text{if } i \text{ is odd}
\end{cases}
\]

where \( d \) is the dimensionality of the embeddings.
\vspace{-5pt}
\section{Conclusion}
\label{sec:conclusion}

In the past, dynamic layer sparsity has not been practical due to small model sizes and incompatible neural architectures, which caused large contributions from each layer and varying internal dimensions.
For these reasons, it has only been possible with techniques like early-exit, which requires expensive specialized training.
Yet, as language models grow in size, each layer contributes less to output, creating opportunities for dynamic layer sparsity.
Following the trends in Figure~\ref{fig:pareto}, modern language models with over one trillion parameters likely have median residual ratios less than 1\%.
And in the future, as vision and multi-modal models catch up to language models, their residual ratios should follow similar scaling trends.

To take advantage of this dynamic sparsity, we propose \textbf{Radial Networks}, which perform token-level routing between layers guided by a trained router module.
These networks can be used in a post-training distillation from sequential networks or trained from scratch to co-learn the router and layer weights.
They enable scaling to larger model sizes by decoupling the number of layers from the dynamic depth of the network, and their design allows for layer reuse.
By varying the compute token by token, they reduce the overall resources needed for generating entire sequences.
Overall, this leads to larger capacity networks with significantly lower compute and serving costs.

{
    \small
    \bibliographystyle{ieeenat_fullname}
    \bibliography{main}
}

\clearpage
\onecolumn

\appendix 
\section{Appendix}
\label{sec:appendix}

\begin{figure*}[h]
  \centering
  \includegraphics[width=\linewidth]{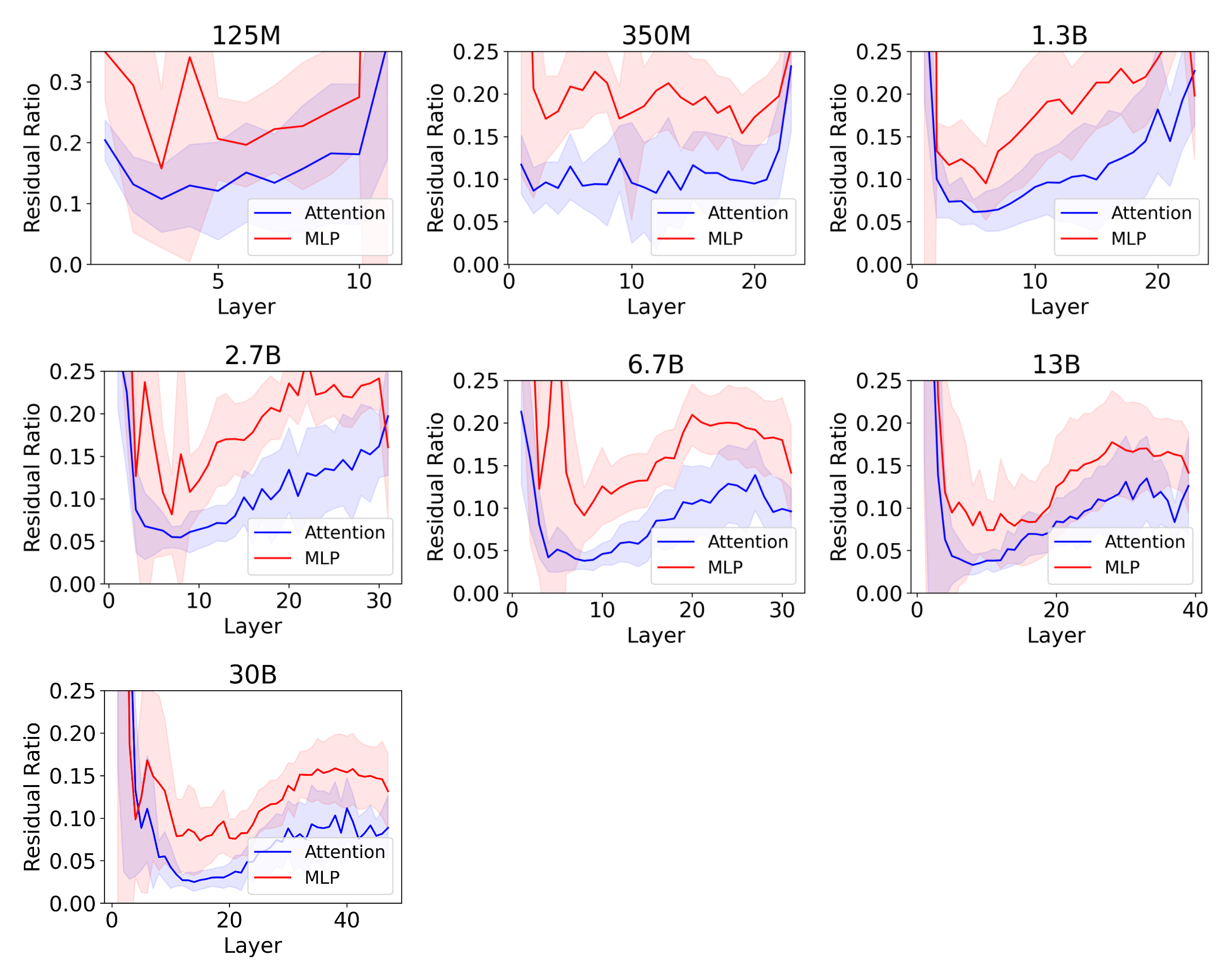}
      \caption{\textbf{OPT Layer Residual Ratio --} In general, the residual ratios decrease with larger model sizes, implying that more layers can be more easily skipped. Across models, there is a U-shaped distribution where the first and last layers have the highest ratios and contribute the most to the model output.}
  \vspace{-5pt}
  \label{fig:opt-layers}
\end{figure*}

\begin{figure*}[h]
  \centering
  \includegraphics[width=\linewidth]{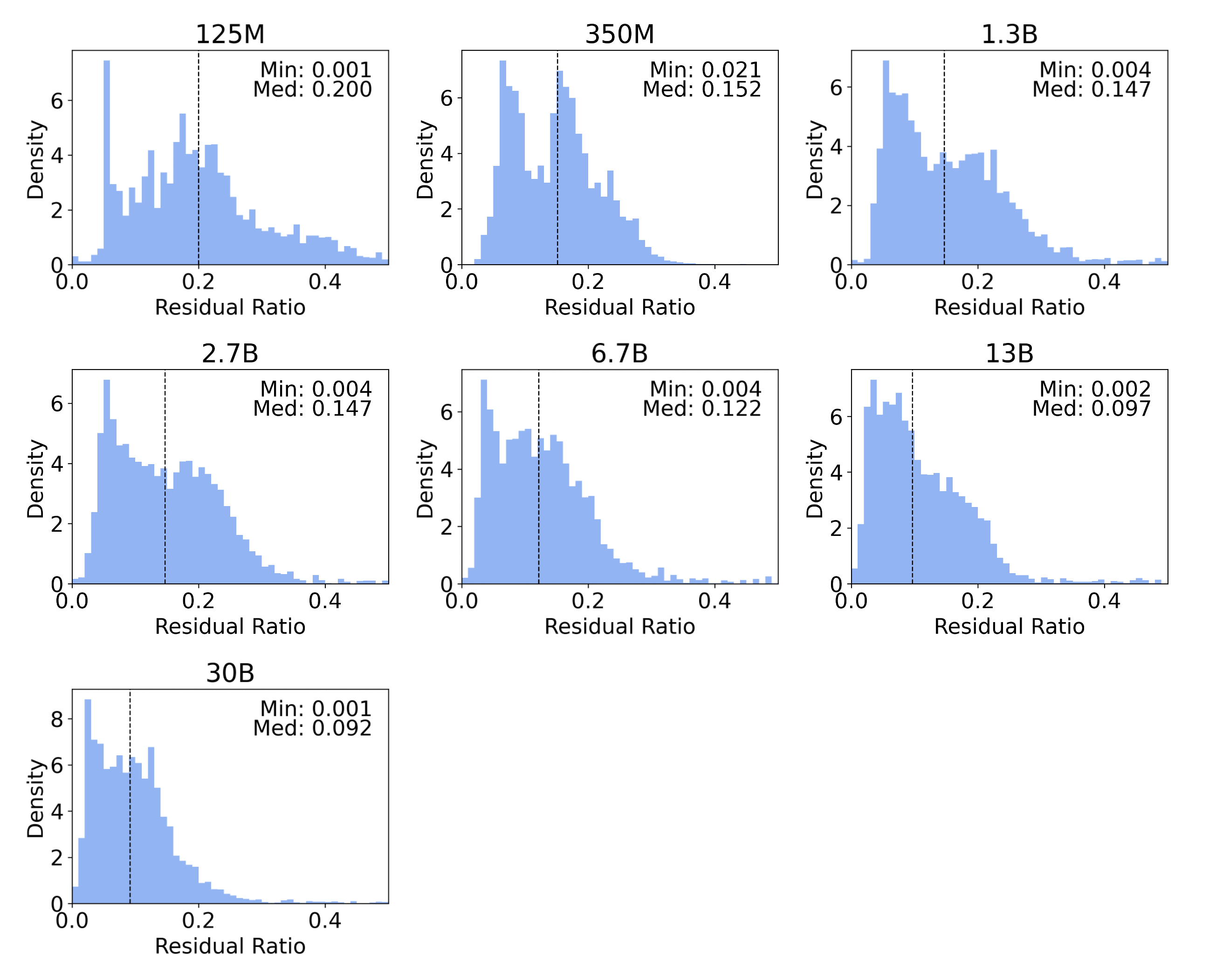}
      \caption{\textbf{OPT Residual Ratios --} As models grow larger, the residual ratios shift lower indicating more dynamic layer sparsity within the model. Each value represents the residual ratio from a single layer and single token. Evaluation is done on batches of sequences with length 256 sampled from the WikiText-2 dataset. The dashed line represents the median value in the distribution.}
  \vspace{-5pt}
  \label{fig:opt-hist}
\end{figure*}

\begin{figure*}[h]
  \centering
  \includegraphics[width=\linewidth]{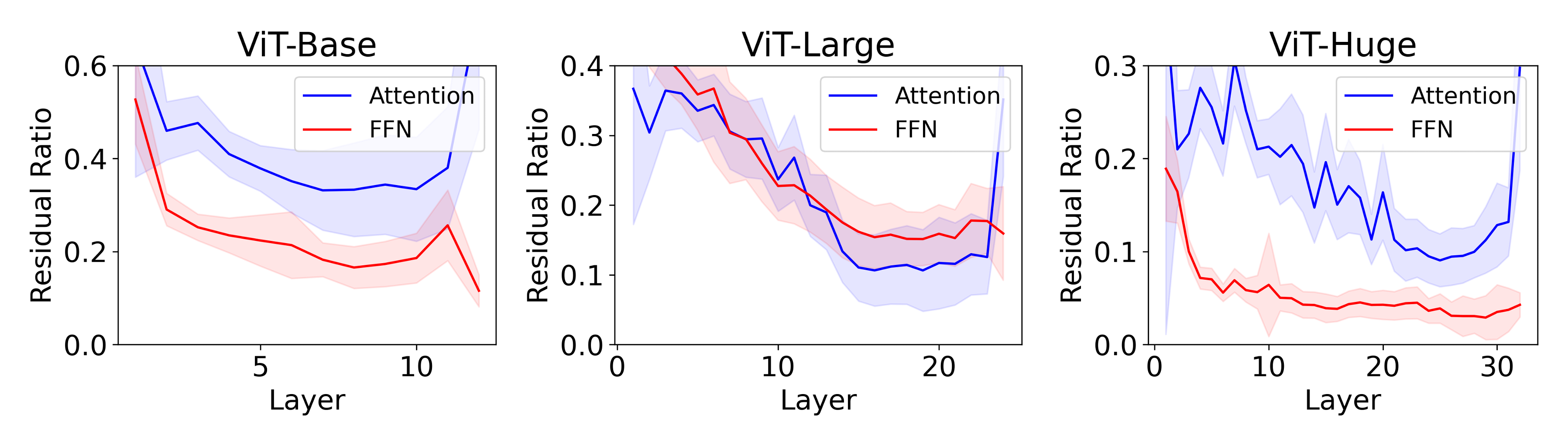}
   \caption{\textbf{ViT Family --}  Current vision transformers are substantially smaller than language models yet demonstrate similar trends with residual ratio. As model size increases, the contributions of each layer decreases.}
    \vspace{-5pt}
  \label{fig:vit-sizes}
\end{figure*}

\end{document}